\documentclass{article}

\usepackage[preprint]{arxiv}

\makeatletter
\renewcommand{\@toptitlebar}{}
\renewcommand{\@bottomtitlebar}{}
\makeatother

\usepackage[utf8]{inputenc} 
\usepackage[T1]{fontenc}    
\usepackage{hyperref}       
\usepackage{url}            
\usepackage{booktabs}       
\usepackage{amsfonts}       
\usepackage{nicefrac}       
\usepackage{microtype}      
\usepackage{xcolor}         
\usepackage{amsmath}
\usepackage{tabularx}
\usepackage{graphicx}     
\usepackage{array}  
\usepackage{colortbl}
\usepackage{wrapfig} 
\usepackage{fontawesome5} 
\usepackage{subcaption}  

\usepackage{multirow}
\usepackage{amsmath}
\usepackage{mathtools}              
\usepackage{enumitem}   
\usepackage{graphicx}               
\usepackage{tabularx}               
\usepackage{booktabs}
\usepackage{amsmath}
\usepackage{stmaryrd}
\usepackage{xcolor}
\usepackage{soul}

\usepackage{graphicx}               
\usepackage{tabularx}               
\usepackage{booktabs}
\usepackage{amsmath}
\usepackage{stmaryrd}
\usepackage{xcolor}
\usepackage{soul}

\newcolumntype{C}{>{\centering\arraybackslash}X}
\usepackage{multirow}               
\usepackage{diagbox}                
\usepackage{hhline}                 
\usepackage{color}                  
\usepackage{amsmath}                
\usepackage{mathtools}              
\usepackage{enumitem} 
\usepackage{booktabs}
\usepackage{extarrows}
\usepackage{makecell}
\usepackage{xparse}
\usepackage{soul}
\usepackage{xcolor}
\usepackage[linesnumbered,ruled,vlined]{algorithm2e}
\usepackage{adjustbox} 

\usepackage{dsfont}

\newcommand{\mat}[1]{\ensuremath{\mathbf{\uppercase{#1}}}} 

\definecolor{RoseQuartzBg}{HTML}{F7CAC9}
\definecolor{ForestGreen}{HTML}{228b22}
\definecolor{RoseQuartz}{HTML}{F5A798}
\definecolor{Serenity}{HTML}{92A8D1}
\definecolor{OrangeRed}{rgb}{1.0, 0.27, 0.0}
\definecolor{Red}{rgb}{1.0, 0.0, 0.0}
\definecolor{forestgreen}{rgb}{0.13, 0.55, 0.13}
\definecolor{Turquoise}{HTML}{0F4C81}
\definecolor{columbiablue}{rgb}{0.61, 0.87, 1.0}
\definecolor{Gray}{gray}{0.9}

\usepackage{xparse}
\usepackage{footmisc}

\NewDocumentCommand{\zhiyang}{ mO{} }{\textcolor{ForestGreen}{\textsuperscript{\textit{zhiyang}}\textsf{\textbf{\small[#1]}}}}
\newcommand{\HFIcon}{\includegraphics[height=1em]{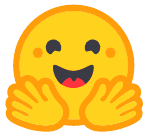}}
\newcommand{\modelname}{BLIP3-o} 
\newcommand{\dataname}{{BLIP3o-60k}} 

\usepackage[most]{tcolorbox}

\tcbset{
  finding style/.style={
    enhanced,
    colback=gray!10,           
    colframe=gray!60,         
    boxrule=0.8pt,             
    arc=2mm,                   
    left=1mm, right=1mm,       
    top=1mm, bottom=1mm,
    before skip=10pt, after skip=10pt,
    fontupper=\normalfont,     
    fonttitle=\normalfont\bfseries, 
  }
}

\newtcolorbox{findingbox}[1][]{finding style,#1}

\title{\modelname: A Family of Fully Open Unified Multimodal Models—Architecture, Training and Dataset}

\author{
\textbf{Jiuhai Chen}\textsuperscript{1,2*} \quad
\textbf{Zhiyang Xu}\textsuperscript{3*} \quad
\textbf{Xichen Pan}\textsuperscript{4*} \quad
\textbf{Yushi Hu}\textsuperscript{5*} \\
\textbf{Can Qin}\textsuperscript{1}, \textbf{Tom Goldstein}\textsuperscript{2}, \textbf{Lifu Huang}\textsuperscript{6}, 
\textbf{Tianyi Zhou}\textsuperscript{2}, \textbf{Saining Xie}\textsuperscript{4}, \textbf{Silvio Savarese}\textsuperscript{1} \\
\textbf{Le Xue}\textsuperscript{1\dag}, \textbf{Caiming Xiong}\textsuperscript{1\ddag}, \textbf{Ran Xu}\textsuperscript{1\ddag} \\
\\
\textsuperscript{1}Salesforce Research \\
\textsuperscript{2}University of Maryland, \quad
\textsuperscript{3}Virginia Tech, \quad
\textsuperscript{4}New York University, \\
\textsuperscript{5}University of Washington , \quad
\textsuperscript{6} UC Davis \\
\\
\textsuperscript{*}Equal Contribution. \quad
\textsuperscript{\dag}Project Lead. \quad
\textsuperscript{\ddag}Corresponding Authors.
}

\begin{document}

\maketitle

\begin{abstract}
Unifying image understanding and generation has gained growing attention in recent research on multimodal models. Although design choices for image understanding have been extensively studied, the optimal model architecture and training recipe for a unified framework with image generation remain underexplored. Motivated by the strong potential of autoregressive and diffusion models for high-quality generation and scalability, we conduct a comprehensive study of their use in unified multimodal settings, with emphasis on image representations, modeling objectives, and training strategies.
Grounded in these investigations, we introduce a novel approach that employs a diffusion transformer to generate semantically rich CLIP image features, in contrast to conventional VAE-based representations. This design yields both higher training efficiency and improved generative quality. Furthermore, we demonstrate that a sequential pretraining strategy for unified models—first training on image understanding and subsequently on image generation—offers practical advantages by preserving image-understanding capability while developing strong image generation ability.
Finally, we carefully curate a high-quality instruction-tuning dataset \dataname{} for image generation by prompting GPT-4o with a diverse set of captions covering various scenes, objects, human gestures, and more.
Building on our innovative model design, training recipe, and datasets, we develop \modelname{}, a suite of state-of-the-art unified multimodal models. \modelname{} achieves superior performance across most of the popular benchmarks spanning both image understanding and generation tasks.
\textit{To facilitate future research, we fully open-source our models, including code, model weights, training scripts, and pretraining and instruction tuning datasets.}
\end{abstract}

\begin{table}[ht]
  \centering
  \begin{tabular}{@{}l l@{}}
    \faGithub\quad Code:   & \url{https://github.com/JiuhaiChen/BLIP3o} \\[4pt]
    \HFIcon\quad Models:   & \url{https://huggingface.co/BLIP3o/BLIP3o-Model} \\[4pt]
    \HFIcon\quad Pretrain Data:     & \url{https://huggingface.co/datasets/BLIP3o/BLIP3o-Pretrain} \\[4pt]
    \HFIcon\quad Instruction Tuning Data:     & \url{https://huggingface.co/datasets/BLIP3o/BLIP3o-60k} \\
  \end{tabular}
\end{table}

\newpage

\tableofcontents 

\newpage

\begin{figure}[!t]
\centering
\includegraphics[width=\linewidth]{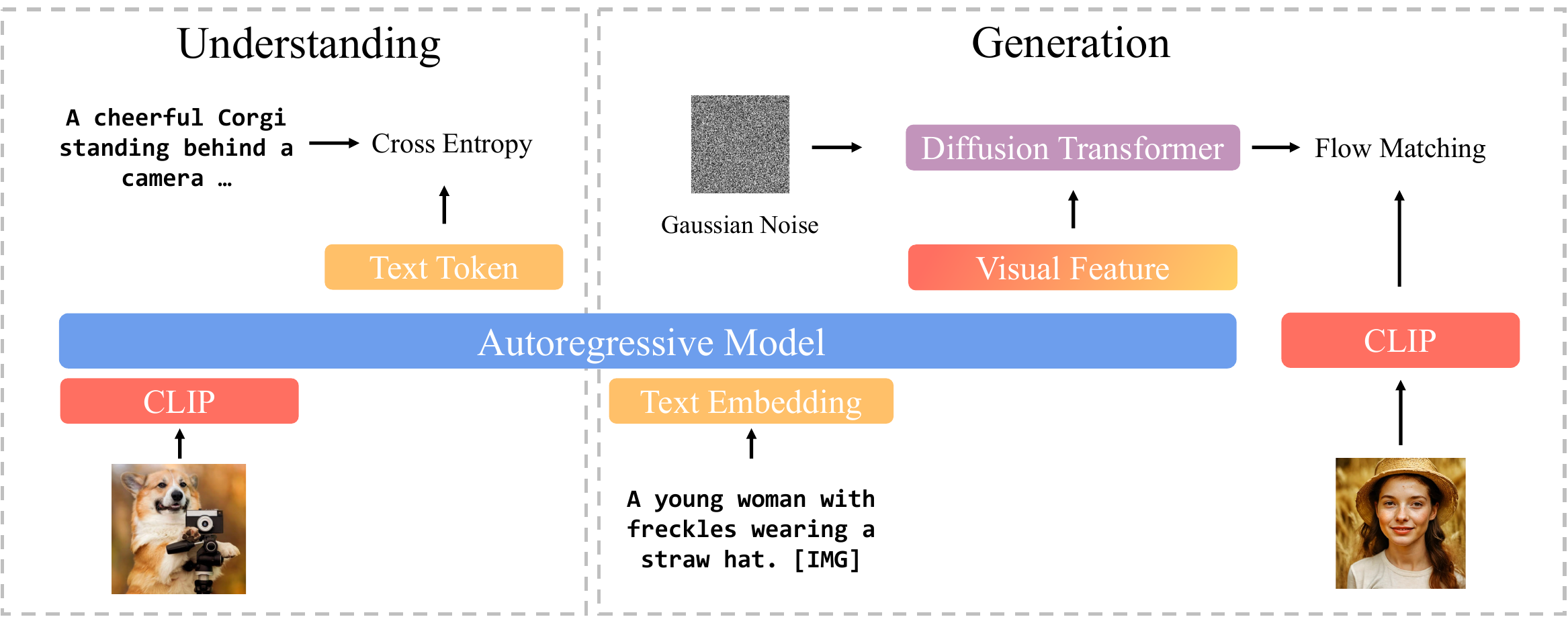}
\vspace{-1em}
\caption{The architecture of \modelname{}. For image understanding part, we use CLIP to encode the image and compute the cross entropy loss between the target text token and predicted text token. For image generation part, autoregressive model first generates a sequence of intermediate visual
features, which are then used as conditioning inputs to a diffusion transformer that
generates CLIP image features to approximate the ground-truth CLIP features. By using CLIP encoder, image understanding and image generation share the same semantic space, effectively unifying these two tasks.
}
\label{fig:main_archc}
\vspace{-1em}
\end{figure}

\section{Introduction}

Recent advances have demonstrated the potential for unified multimodal representation learning that supports both image understanding and image generation within a single model~\cite{ge2024seed,sun2024generative,xie2024show,wu2024janus,chen2025janus,tong2024metamorph,pan2025transfer}. In this field, despite extensive studies on image understanding, the optimal architecture and training strategy for image generation remain underexplored. The previous debate revolves around two approaches: the first approach quantizes continuous visual features into discrete tokens and models them as categorical distribution~\cite{team2024chameleon,wang2024emu3,ma2025token}; the second approach generates intermediate visual features or latent representations via the autoregressive model and then conditions on these visual features to generate images through the diffusion model~\cite{tong2024metamorph,pan2025transfer}. 
The recently released GPT-4o image generation~\cite{gpt4o} was implied to adopt a hybrid architecture with autoregressive and diffusion models following the second approach~\cite{gpt4o,yan2025gpt}.
Therefore, we were inspired to present a systematic study of design choices in a similar way. Specifically, our investigation focuses on three key design axes:
(1) \textbf{image representations} - whether to encode the images into low-level pixel features (e.g., from VAE-based encoders) or high-level semantic features (e.g., from CLIP image encoders);
(2) \textbf{training objectives} - Mean Squared Error (MSE) versus Flow Matching~\cite{lipman2022flow,RectifiedFlow}, and what their impacts on training efficiency and generation quality;
(3) \textbf{training strategies} - joint multitask training on image understanding and generation like Metamorph~\cite{tong2024metamorph} or sequential training like LMFusion~\cite{shi2024llamafusion} and MetaQuery~\cite{pan2025transfer}, where the model is first trained for understanding and then extended for generation.

Our findings reveal that CLIP image features offer more compact and informative representations than VAE features, resulting in both faster training and higher image generation quality. Flow matching loss proves to be more effective than MSE loss, enabling more diverse image sampling and yielding better image quality.
Furthermore, we find that a sequential training strategy—first training the autoregressive model on image understanding tasks, then freezing it during training on image generation—achieves the best overall performance.
Based on these findings, we develop \modelname{}, a herd of state-of-the-art unified multimodal models. \modelname{} leverages diffusion transformer and flow matching on CLIP features (Figure~\ref{fig:main_archc}) and is sequentially trained on image understanding and image generation tasks. 
To further improve visual aesthetic and instruction following abilities, we carefully curate a 60k high-quality instruction-tuning dataset \dataname{} for image generation, by prompting GPT-4o with a diverse set of prompts spanning scenes, objects, human gestures and more. We observe that supervised instruction tuning on \dataname{} significantly enhances the alignment of \modelname{} with human preference and improves the aesthetic quality.

\begin{figure*}[!t]
  \centering
  \begin{adjustbox}{center}
    \includegraphics[width=1.15\textwidth]{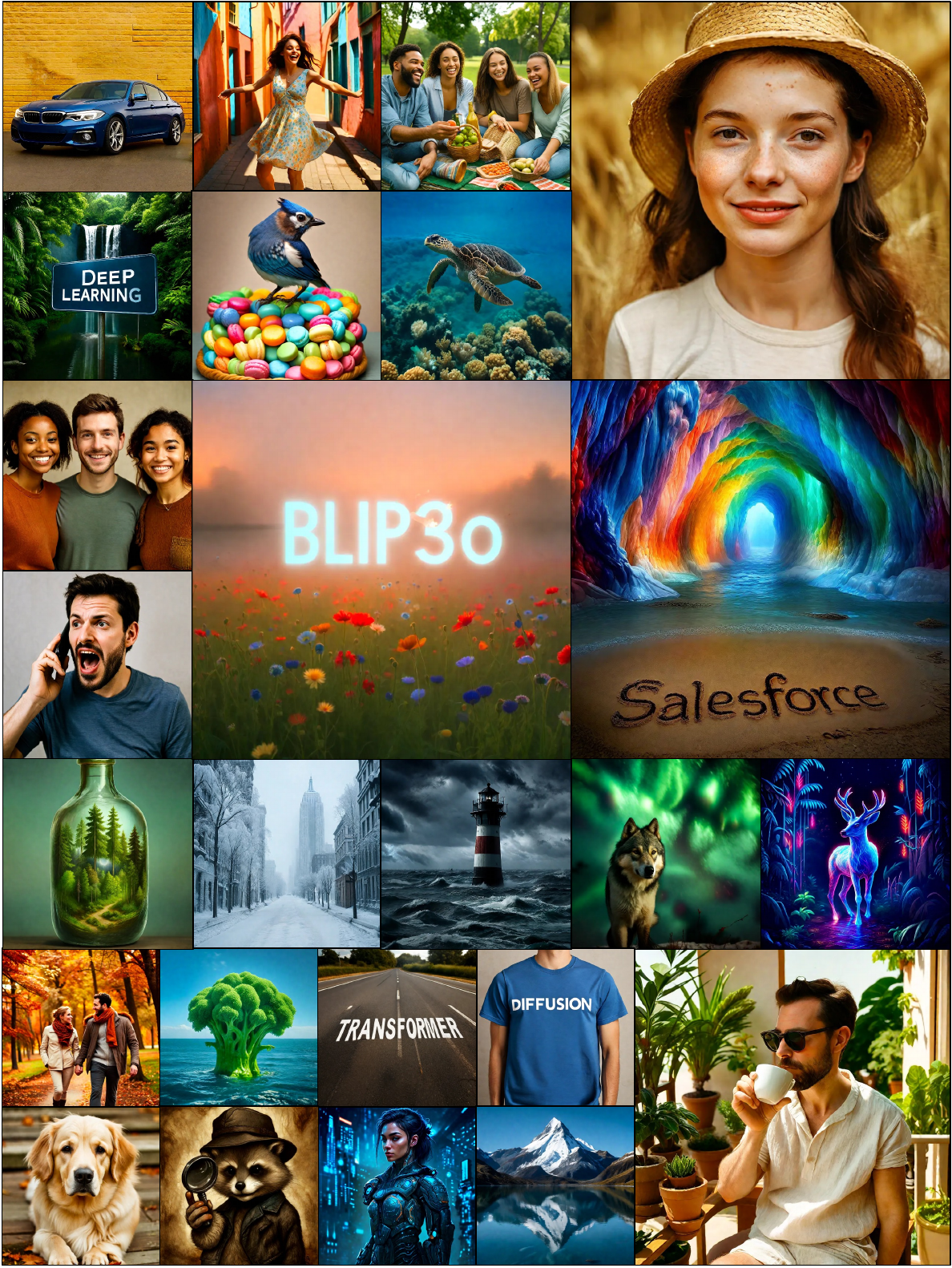}
  \end{adjustbox}
  \caption{
Visualization results of \modelname{} 8B at 1024×1024 resolution.
}
  \label{fig:human_fig}
\end{figure*}

In our experiments,  \modelname{} achieves superior performance across most of the popular benchmarks for image understanding and image generation, with the 8B model scoring 1682.6 on MME-P, 50.6 on MMMU and 0.84 on GenEval. \textit{To support further research and keep the mission of open-source foundation model research like BLIP-3~\cite{blip3}, we fully open-source our models including model weights, code, pretraining and instruction-tuning datasets, and evaluation pipelines. We hope that our work will support the research community and drive continued progress in the unified multimodal domain.}


\section{Unified Multimodal for Image Generation and Understanding}

\subsection{Motivation}

The development of unified multimodal architectures that jointly support both image understanding and generation has emerged as a promising direction in recent research. Models such as Janus~\cite{chen2025janus}, Show‑o~\cite{xie2024show}, MetaMorph~\cite{tong2024metamorph}, Janus-Pro~\cite{chen2025janus}, and LMFusion~\cite{shi2024llamafusion} exemplify early efforts to bridge image understanding and generation within a single framework. More recently, OpenAI’s GPT-4o~\cite{gpt4o} has further sparked interest in this paradigm by demonstrating impressive capabilities in both high-quality image generation and strong multimodal understanding.
Despite this growing interest, the underlying design principles and training strategies that enable such unified capabilities remain underexplored. This work aims to systematically investigate and advance the development of unified models, and we begin by clearly presenting the key motivations for building unified multimodal model.


\paragraph{Reasoning and Instruction Following} Integrating image generation capabilities into autoregressive models such as Multimodal Large Language Models (MLLMs) offers the potential to inherit their pretrained knowledge, reasoning capability and instruction following ability. For example, our model is able to interpret prompts, such as ``An animal with a long nose”, without requiring prompt rewriting. This demonstrates a level of reasoning capability and world knowledge that traditional image generation models struggle to achieve. Beyond reasoning, the instruction-following capabilities of MLLMs are expected to carry over to the image generation process when incorporated into a unified architecture.
\paragraph{In-context Learning} Unified models that jointly support image understanding and generation naturally facilitate in-context learning capabilities. In such models, previously generated multimodal outputs can serve as context for subsequent generation, enabling seamless support for iterative image editing, visual dialogue, and step-by-step visual reasoning. This eliminates the need for mode switching or reliance on external processing pipelines, allowing the model to maintain coherence and task continuity.



\paragraph{Towards Multimodal AGI} As artificial intelligence advances toward artificial general intelligence (AGI), future systems need to go beyond text-based capabilities to seamlessly perceive, interpret, and generate multimodal content. Achieving this requires a shift from text-only architectures to unified multimodal architectures that can reason and generate across diverse modalities. Such models are essential for building general-purpose intelligence capable of engaging with the world in a holistic, human-like manner.

Driven by these motivations, we explore the development of a unified model that jointly supports image understanding and generation tasks in the following sections.

\subsection{Combining Autoregressive and Diffusion Models}
Recent OpenAI's GPT-4o~\cite{gpt4o} has demonstrated state-of-the-art performance in image understanding, generation and editing tasks. Emerging hypotheses of its architecture~\cite{yan2025gpt} suggest a hybrid pipeline structured as:
\[
\textbf{Tokens} \;\longrightarrow\; \textbf{[Autoregressive Model]} \;\longrightarrow\; \textbf{[Diffusion Model]} \;\longrightarrow\; \textbf{Image Pixels}
\]
indicating that autoregressive and diffusion models may be jointly leveraged to combine the strengths of both modules. Motivated by this hybrid design, we adopt an autoregressive + diffusion framework in our study. However, the optimal architecture in this framework remains unclear.
The autoregressive model produces continuous intermediate visual features intended to approximate ground-truth image representations, raising two key questions. First, what should serve as the ground-truth embeddings: should we use a VAE or CLIP to encode images into continuous features? Second, once the autoregressive model generates visual features, how do we optimally align them with the ground-truth image features, or more generally, how should we model the distribution of these continuous visual features: via a simple MSE loss, or by employing a diffusion-based approach?
Thus, we conduct a comprehensive exploration of various design choices in the following section.




\section{Image Generation in Unified Multimodal}
In this section, we discuss the design choices involved in building the image generation model within a unified multimodal framework. We begin by exploring how images can be represented as continuous embeddings through encoder–decoder architectures, which play a foundational role in learning efficiency and generation quality.

\subsection{Image Encoding and Reconstruction}


Image generation typically begins by encoding an image into a continuous latent embedding using an encoder, followed by a decoder that reconstructs the image from this latent embedding. This encoding-decoding pipeline can effectively reduce the dimensionality of the input space in image generation, facilitating efficient training. In the following, we discuss two widely used encoder–decoder paradigms.

\paragraph{Variational Autoencoders}
Variational Autoencoders (VAEs)~\cite{kingma2013auto, rezende2014stochastic} are a class of generative models that learn to encode images into a structured, continuous latent space. The encoder approximates the posterior distribution over the latent variables given the input image, while the decoder reconstructs the image from samples drawn from this latent distribution.
Latent diffusion models build on this framework by learning to model the distribution of compressed latent representations, rather than raw image pixels. By operating in the VAE latent space, these models significantly reduce the dimensionality of the output space, thereby lowering computational costs and enabling more efficient training. After the denoising steps, the VAE decoder maps the generated latent embeddings into raw image pixels.

\paragraph{CLIP Encoder with Diffusion Decoder}
CLIP~\cite{radford2021learning} models have become foundational encoders for image understanding tasks~\cite{liu2023visual}, owing to its strong ability to extract rich, high-level semantic features from images through contrastive training on large-scale image–text pairs. However, leveraging these features for image generation remains a non-trivial challenge, as CLIP was not originally designed for reconstruction tasks.
Emu2~\cite{sun2024generative} presents a practical solution by pairing a CLIP-based encoder with a diffusion-based decoder. Specifically, it uses EVA-CLIP to encode images into continuous visual embeddings and reconstructs them via a diffusion model initialized from SDXL-base~\cite{podell2023sdxl}. During training, the diffusion decoder is fine-tuned to use the visual embeddings from EVA-CLIP as conditions to recover the original image from Gaussian noise, while the EVA-CLIP remains frozen. This process effectively combines the CLIP and diffusion models into an image autoencoder: the CLIP encoder compresses an image into semantically rich latent embeddings, and the diffusion-based decoder reconstructs the image from these embeddings. Notably, although the decoder is based on diffusion architecture, it is trained with a reconstruction loss rather than probabilistic sampling objectives. Consequently, during inference, the model performs deterministic reconstruction.

 
\paragraph{Discussion}
These two encoder–decoder architectures, i.e., VAEs and CLIP-Diffusion, represent distinct paradigms for image encoding and reconstruction, each offering specific advantages and trade-offs. VAEs encode the image into low-level pixel features and offer better reconstruction quality. Furthermore, VAEs are widely available as off-the-shelf models and can be integrated directly into image generation training pipelines. In contrast, CLIP-Diffusion requires additional training to adapt the diffusion models to various CLIP encoders.
However, CLIP-Diffusion architectures offer significant benefits in terms of image compression ratio. For example, in both Emu2~\cite{sun2024generative} and our experiments, each image regardless of its resolution can be encoded into a fixed length of 64 continuous vectors, providing both compact and semantically rich latent embeddings. By contrast, VAE-based encoders tend to produce a longer sequence of latent embeddings for higher-resolution inputs, which increases the computational burden in the training procedure. 

\subsection{Modeling Latent Image Representation}

After obtaining continuous image embeddings, we proceed to model them using autoregressive architectures. Given a user prompt (e.g., \textit{``A young woman with
freckles wearing a
straw hat.''}), we first encode the prompt into a sequence of embedding vectors \(\mathbf{C}\) using the autoregressive model's input embedding layer, and append a learnable query vector \(\mathbf{Q}\) to \(\mathbf{C}\), where \(\mathbf{Q}\) is randomly initialized and optimized during training.
As the combined sequence \([\mathbf{C}; \mathbf{Q}]\) is processed through the autoregressive transformer, \(\mathbf{Q}\) learns to attend to and extract relevant semantic information from the prompt \(\mathbf{C}\). The resulting \(\mathbf{Q}\) is interpreted as the intermediate visual features or latent representation generated by the autoregressive model, and is trained to approximate the ground-truth image feature \( \mathbf{X} \) (obtained from VAE or CLIP).
In the following, we introduce two training objectives: Mean Squared Error (MSE) and Flow Matching, for learning to align \(\mathbf{Q}\) with the ground-truth image embedding \(\mathbf{X}\).

\paragraph{MSE Loss} The Mean Squared Error (MSE) loss is a straightforward and widely used objective for learning continuous image embeddings~\cite{ge2024seed, sun2024generative}. Given the predicted visual features \( \mathbf{Q} \) produced by the autoregressive model and the ground-truth image features \( \mathbf{X} \), we first apply a learnable linear projection to align the dimensionality of \( \mathbf{Q} \) with that of \( \mathbf{X} \). The MSE loss is then formulated as:
\[
\mathcal{L}_{\text{MSE}} = \left\| \mathbf{X} - \mathbf{W}\mathbf{Q} \right\|_2^2,
\]
where \( \mathbf{W} \) denotes the learnable projection matrix.

\paragraph{Flow Matching} 
Note that using MSE loss only aligns the predicted image features \( \mathbf{Q} \) with the mean value of the target distribution. An ideal training objective would model the probability distribution of continuous image representation.
We propose to use flow matching \cite{FlowMatching}, a diffusion framework that can sample from a target continuous distribution by iterative transporting samples from a prior distribution (e.g., Gaussian).
Given a ground-truth image feature $\mat{X}_1$ and the condition $\mathbf{Q}$ encoded by an autoregressive model, at each training step, we sample a timestep $t\sim\mathcal{U}(0,1)$, and noise $\mat{X}_0\sim\mathcal{N}(0,1)$.  Then diffusion transformer learns to predict the velocity $\mat{V}_t=\frac{d\mat{X}_t}{dt}$ at the timestep $t$ conditioned on $\mathbf{Q}$, in the direction of $\mat{X}_1$.
Following previous work~\cite{RectifiedFlow}, we compute $\mat{X}_t$ by a simple linear interpolation between $\mat{X}_0$ and $\mat{X}_1$:
\[
\mat{X}_t = t\mat{X}_t + (1-t)\mat{X}_0
,\]
and the analytical solution of $\mat{V}_t$ can be expressed as:
\[
\mathbf{V}_t = \frac{d\mathbf{X}_t}{dt} 
= \mathbf{X}_t - \mathbf{X}_0
.\]
Finally, the training objective is defined as:
\[
    \mathcal{L}_{\text{Flow}}(\theta) = \mathbb{E}_{(\mat{X}_1,\mathbf{Q})\sim\mathcal{D}, t\sim\mathcal{U}(0,1),\mat{X}_0\sim\mathcal{N}(0,1)} \left[ \|\mat{V}_\theta(\mat{X}_t,\mathbf{Q},t) - \mat{V}_t\|^2 \right],
\]
where $\theta$ is the diffusion transformer's parameters, and $\mat{V}_\theta(\mat{X}_t,\mathbf{Q},t)$ denotes the predicted velocity based on an instance $(\mat{X}_1,\mathbf{Q})$, timestep $t$, and noise $\mat{X}_0$.

\paragraph{Discussion}
Unlike discrete tokens, which inherently support sampling-based strategies for exploring diverse generation paths, continuous representations lack this property. Specifically, under an MSE-based training objective, the predicted visual features \( \mathbf{Q} \) becomes nearly deterministic for a given prompt. As a result, the output images, regardless of whether the visual decoder is based on VAEs or CLIP + Diffusion architectures, remain almost identical across multiple inference runs. This determinism highlights a key limitation of the MSE objective: it constrains the model to produce a single, fixed output for each prompt, thereby limiting generation diversity.

In contrast, the flow matching framework enables the model to inherit the stochasticity of the diffusion process. This allows the model to generate diverse image samples conditioned on the same prompt, facilitating broader exploration of the output space. However, this flexibility comes at the cost of increased model complexity. Flow matching introduces additional learnable parameters compared to MSE. In our implementation, we use a diffusion transformer (DiT), and empirically find that scaling its capacity yields significant performance improvements.


\subsection{Design Choices}
\label{design_choice}
\begin{figure}[h!]
\centering
\includegraphics[width=\linewidth]{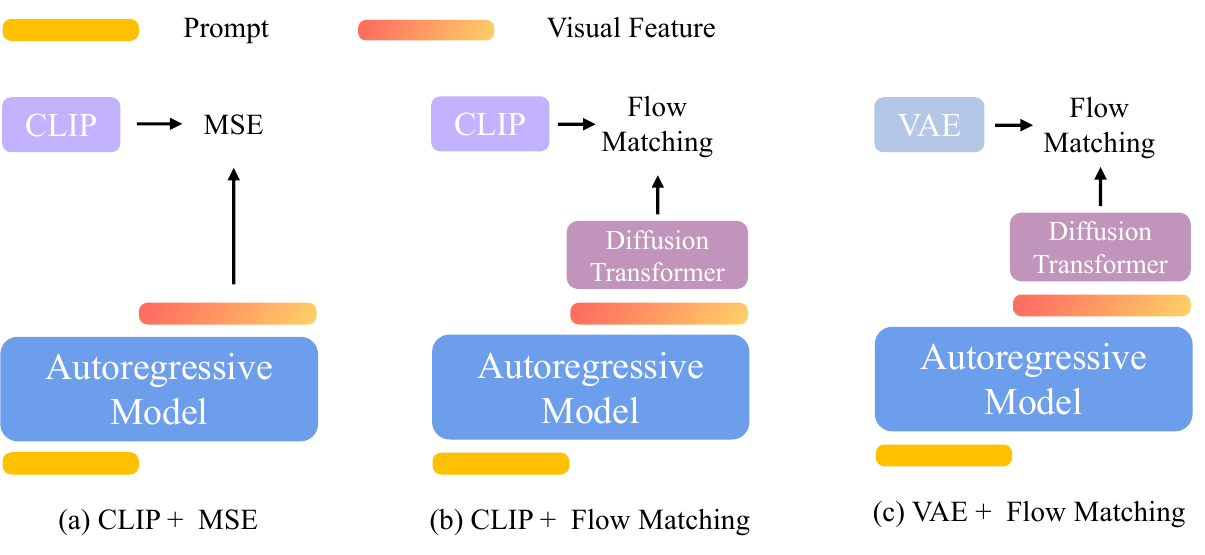}
\vspace{-1em}
\caption{Three design choices for image generation in unified multimodal model. All designs use a \textbf{Autoregressive + Diffusion} framework but vary in their image generation components. For the flow matching loss, we keep the autoregressive model frozen and only fine-tune the image generation module to preserve the model’s language capabilities. }
\label{fig:design_choice}
\vspace{-1em}
\end{figure}

The combination of different image encoder–decoder architectures and training objectives gives rise to a range of design choices for image generation models. These design choices, illustrated in Figure~\ref{fig:design_choice}, significantly influence both the quality and controllability of the generated images. In this section, we summarize and analyze the trade-offs introduced by different encoder types (e.g., VAEs vs. CLIP encoders) and loss functions (e.g., MSE vs. Flow Matching).

\paragraph{CLIP + MSE}  
Following Emu2~\cite{sun2024generative}, Seed-X~\cite{ge2024seed} and Metamorph~\cite{tong2024metamorph}, we use CLIP to encode images into 64 fixed-length semantic-rich visual embeddings. The autoregressive model is trained to minimize the Mean Squared Error (MSE) loss between the predicted visual features \(\mathbf{Q}\) and the ground-truth CLIP embedding \(\mathbf{X}\), as illustrated in Figure~\ref{fig:design_choice}(a). 
During inference, given a text prompt \(\mathbf{C}\), the autoregressive model predicts the latent visual features \(\mathbf{Q}\), which is subsequently passed to a diffusion-based visual decoder to reconstruct the real image.


\paragraph{CLIP + Flow Matching}  
As an alternative to MSE loss, we employ flow matching loss to train the model to predict ground-truth CLIP embeddings, as illustrated in Figure~\ref{fig:design_choice}(b). 
Given a prompt \(\mathbf{C}\), the autoregressive model generates a sequence of visual features \(\mathbf{Q}\). These features are used as conditions to guide the diffusion process, yielding a predicted CLIP embedding to approximate the ground-truth CLIP features. In essence, the inference pipeline involves two diffusion stages: 
the first uses the conditioning visual features \(\mathbf{Q}\) to iteratively denoise into CLIP embeddings. And the second converts these CLIP embeddings into real images by diffusion-based visual decoder. This approach enables stochastic sampling at the first stage, allowing for greater diversity in image generation.

\paragraph{VAE + Flow Matching} 

We can also use flow matching loss to predict the ground truth VAE features seen in Figure~\ref{fig:design_choice}(c), which is similar to MetaQuery~\cite{pan2025transfer}. At inference time, given a prompt \(\mathbf{C}\), the autoregressive model produces visual features \(\mathbf{Q}\). Then, conditioning on \(\mathbf{Q}\) and iteratively removing noise at each step, the real images are generated by the VAE decoder.


\paragraph{VAE + MSE} 

Because our focus is on autoregressive + diffusion framework, we exclude VAE + MSE approaches, as they do not incorporate any diffusion module.

\paragraph{Implementation Details}

To compare various design choices, we use Llama-3.2-1B-Instruct as autoregressive model. Our training data consists of CC12M \cite{changpinyo2021conceptual}, SA-1B \cite{kirillov2023segment}, and JourneyDB \cite{JourneyDB}, amounting to approximately 25 million samples. For CC12M and SA-1B, we utilize the detailed captions generated by LLaVA, while for JourneyDB we use the original captions. 
The detailed description of image generation architecture using flow matching loss is provided in Section~\ref{sec:model_arc}.



\begin{figure}[h!]
\centering
\includegraphics[width=\linewidth]{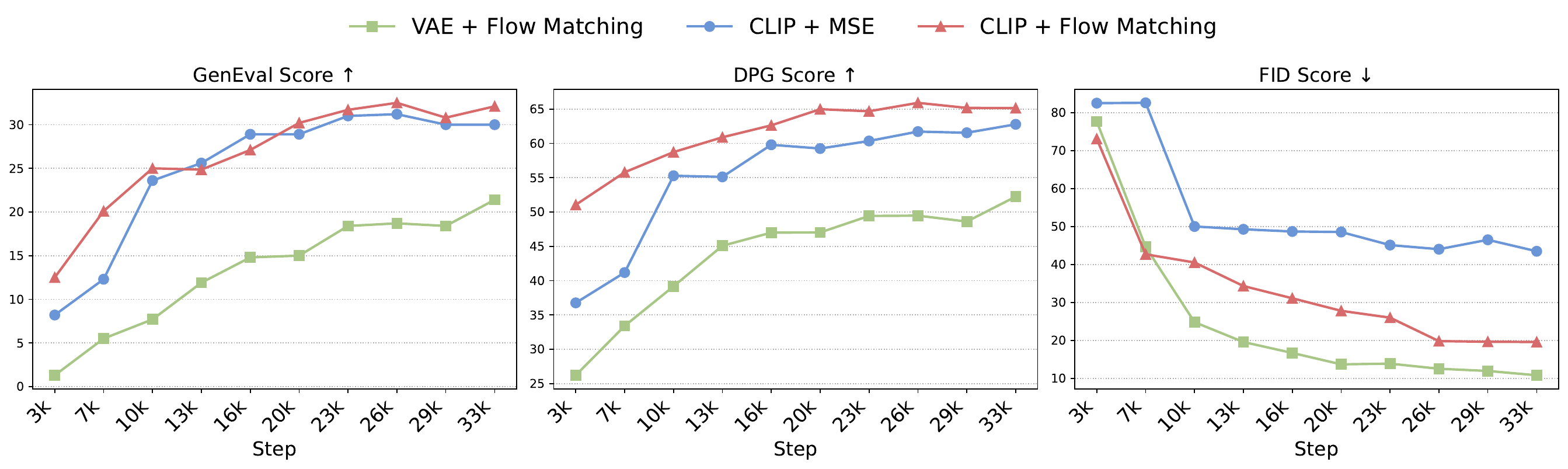}
\vspace{-1em}
\caption{Comparison of different design choices.}
\label{fig:ablation_study}
\vspace{-1em}
\end{figure}

\paragraph{Results}
We report the FID score \cite{heusel2017gans} on MJHQ-30k \cite{li2024playground} for visual aesthetic quality, along with GenEval \cite{ghosh2023geneval} and DPG-Bench \cite{hu2024ella} metrics for evaluating prompt alignment. 
We plot the results for each design choice at approximately every 3,200 training steps.
Figure~\ref{fig:ablation_study} shows that CLIP + Flow Matching achieves the best prompt alignment scores on both GenEval and DPG-Bench, while VAE + Flow Matching produces the lowest (best) FID, indicating superior aesthetic quality. However, FID has inherent limitations: it quantifies stylistic deviation from the target image distribution and often overlooks true generative quality and prompt alignment.  In fact, our FID evaluation of GPT-4o on the MJHQ-30k dataset produced a score of around 30.0, underscoring that FID can be misleading in the image generation evaluation. In general, our experiments demonstrate CLIP + Flow Matching as the most effective design choice. 

\paragraph{Discussion} 

In this section, we present a comprehensive evaluation of various design choices for image generation within a unified multimodal framework. Our results clearly show that CLIP's features produce more compact and semantically rich representations than VAE features, yielding higher training efficiency. Autoregressive models more effectively learn these semantic-level features compared to pixel-level features.
Furthermore, flow matching proves to be a more effective training objective for modeling the image distribution, resulting in greater sample diversity and enhanced visual quality.

\begin{findingbox}[title={Finding 1}]
When integrating image generation into a unified model, autoregressive models more effectively learn the semantic-level features (CLIP) compared to pixel-level features (VAE).
\end{findingbox}

\begin{findingbox}[title={Finding 2}]
Adopting flow matching as the training objective better captures the underlying image distribution, resulting in greater sample diversity and enhanced visual quality.
\end{findingbox}

\section{Training Strategies for Unified Multimodal}

\begin{figure}[h!]
\centering
\includegraphics[width=\linewidth]{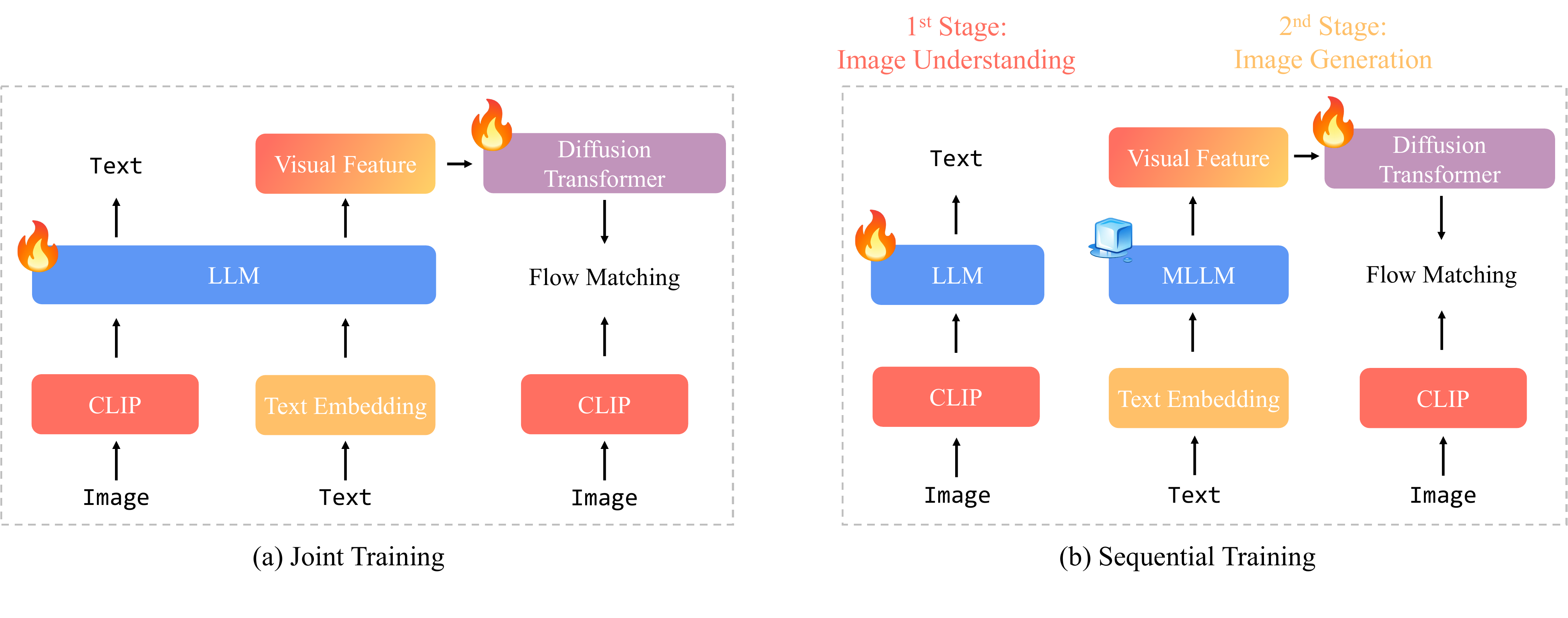}
\caption{
Joint Training vs. Sequential Training: Joint training performs multitask learning by mixing image‑understanding and image‑generation data, updating both the autoregressive backbone and the generation module simultaneously. Sequential training separates the process: first, the model is trained only on image‑understanding tasks; then the autoregressive backbone is frozen and only the image‑generation module is trained in a second stage.
}
\label{fig:training_stra}
\end{figure}

Building on our image generation study, the next step is to develop a unified model that can perform both image understanding and image generation. We use CLIP + Flow Matching for the image generation module. 
Since image understanding also operates in CLIP’s embedding space, we align both tasks within the same semantic space, enabling their unification.
In this context, we discuss two training strategies to achieve this integration.

\subsection{Joint Training Versus Sequential Training} 

\paragraph{Joint Training} Joint training of image understanding and image generation has become a common practice in recent works such as Metamorph \cite{tong2024metamorph}, Janus-Pro \cite{chen2025janus}, and Show-o \cite{xie2024show}. Although these methods adopt different architectures for image generation, all perform multitask learning by mixing data for image generation and understanding.

\paragraph{Sequential Training} 
Instead of training image understanding and generation together, we follow a two‐stage approach. In the first stage, we train only the image understanding module. In the second stage, we freeze the MLLM backbone, and train only the image generation module like LMFusion~\cite{shi2024llamafusion} and MetaQuery~\cite{pan2025transfer}.

\subsection{Discussion}

In joint training setting, although image understanding and generation tasks possibly benefit each other as demonstrated by Metamorph~\cite{tong2024metamorph}, two critical factors influence their synergistic effect: (i) the total data size and (ii) the data ratio between image understanding and generation data. 
In contrast, sequential training offers greater flexibility: It lets us freeze the autoregressive backbone and maintain the image understanding capability. We can dedicate all training capacity to image generation, avoiding any inter-task effects in joint training.  Also motivated by LMFusion~\cite{shi2024llamafusion} and MetaQuery~\cite{pan2025transfer}, we will choose sequential training to construct our unified multimodal model and defer joint training to future work.



\section{\modelname: Our State-of-the-Art Unified Multimodal}





Based on our findings, we adopt CLIP + Flow Matching and sequential training to develop our own state-of-the-art unified multimodal model \modelname.


\subsection{Model Architecture}
\label{sec:model_arc}
We develop two different size models: 8B parameter model trained on proprietary data and 4B parameter model using \textbf{only open source data}. Given the existence of strong open source image understanding models, such as Qwen 2.5 VL~\cite{bai2025qwen2}, we skip image understanding training stage and build our image generation module directly on Qwen 2.5 VL. In the 8B model, we freeze the Qwen2.5-VL-7B-Instruct backbone and train the diffusion transformers, totaling 1.4 B trainable parameters. The 4B model follows the same image generation architecture but uses Qwen2.5-VL-3B-Instruct as backbone.
\paragraph{Diffusion Transformer Architecture} We leverage the architecture of the Lumina-Next model~\cite{zhuo2024lumina} for our DiT. The Lumina-Next model is built on the improved Next-DiT architecture, a scalable and efficient diffusion transformer designed for text-to-image and general multimodal generation. It introduces 3D Rotary Position Embedding to encode spatial-temporal structure across time, height, and width without relying on learnable position tokens. Each transformer block employs sandwich normalization (RMSNorm before and after attention/MLP) and Grouped-Query Attention to enhance stability and reduce computation. Based on empirical results, this architecture achieves fast, high-quality generation.

\subsection{Training Recipe}
\paragraph{Stage 1:  Pretraining for Image Generation}
For 8B model, we combine around 25 million open-source data (CC12M~\cite{changpinyo2021conceptual}, SA-1B~\cite{kirillov2023segment}, and JourneyDB~\cite{JourneyDB}) with an additional 30 million proprietary images. All image captions are generated by Qwen2.5-VL-7B-Instruct, yielding detailed descriptions with an average length of 120 tokens. To improve generalization to varying prompt lengths, we also include around 10\% (6 million) shorter captions with around 20 tokens from CC12M~\cite{changpinyo2021conceptual}. Each image–caption pair is formatted with the prompt:
``Please generate an image based on the following caption: <caption>``.
For the fully open-source 4B model, we use 25 million publicly available images, from CC12M~\cite{changpinyo2021conceptual}, SA-1B~\cite{kirillov2023segment}, and JourneyDB~\cite{JourneyDB}, each paired with the same detailed captions. We also mix in around 10\% (3 million) short captions sourced from CC12M~\cite{changpinyo2021conceptual}. \textbf{To support the research community, we release 25 million detailed captions and 3 million short captions.}

\paragraph{Stage 2: Instruction Tuning for Image Generation}
After the image generation pre-training stage, we observe several weaknesses in the model:
\begin{itemize}
  \item Generate complex human gestures, e.g.\ One person is nocking an arrow.
  \item Generate common objects, such as various fruits and vegetables.
  \item Generate landmarks, for example, Golden Gate Bridge.
  \item Generate simple text, e.g.\ The word  `Salesforce' written on a street surface.
\end{itemize}

Although these categories were intended to be covered during pretraining, the limited size of our pretraining corpus meant they weren’t adequately addressed. To remedy this, we perform instruction tuning focused specifically on these domains. For each category, we prompt GPT-4o to generate roughly 10k prompt–image pairs, creating a targeted dataset that improves the model’s ability to handle these cases.
To improve visual aesthetics quality, we also expand our data with prompts drawn from JourneyDB~\cite{JourneyDB} and DALL·E 3. This process yields a curated collection of approximately 60k high quality prompt–image pairs. We also release this 60k instruction tuning dataset.

\subsection{Results}

For baseline comparison, we include the following unified models: EMU2 Chat~\cite{sun2024generative}, Chameleon~\cite{team2024chameleon}, Seed‑X~\cite{ge2024seed}, VILA‑U~\cite{wu2024vila}, LMfusion~\cite{shi2024llamafusion}, Show‑o~\cite{xie2024show}, EMU3~\cite{wang2024emu3}, MetaMorph~\cite{tong2024metamorph}, TokenFlow~\cite{qu2024tokenflow}, Janus~\cite{wu2024janus}, and Janus‑Pro~\cite{chen2025janus}.

\paragraph{Image Understanding}
In the image understanding task, we evaluate the benchmark performance on VQAv2~\cite{goyal2017making}, MMBench~\cite{liu2023mmbench}, 
SeedBench~\citep{li2023seed},
MM‑Vet~\cite{yu2023mm}, MME‑Perception and MME‑Cognition~\cite{fu2024mme}, MMMU~\cite{yue2024mmmu}, TextVQA~\cite{singh2019towards}, and RealWorldQA~\cite{grok15v}. As shown in Table \ref{tab:und_results}, our \modelname{} 8B achieves the best performance in most benchmarks.

\begin{table*}[h!]
    \centering
    \begin{subtable}[t]{\textwidth}
        \centering
        \fontsize{7pt}{8.8pt}\selectfont
        \setlength\tabcolsep{6pt}
        \renewcommand{\arraystretch}{1.1}
        \resizebox{\textwidth}{!}{
        \begin{tabular*}{\linewidth}{l|ccccccccc}
         Model & VQAv2  & MMBench & SEED   & MM-Vet & MME-P & MME-C  & MMMU & RWQA & TEXTVQA  \\
         \hline
        EMU2 Chat 34B & -  & - & 62.8  & 48.5 & - & -  & 34.1 & - & 66.6  \\
        Chameleon 7B & -  & 19.8 & 27.2 & 8.3 & 202.7 & -  & 22.4 & 39.0 & 0.0 \\
        Chameleon 34B & -  & 32.7 & - & 9.7 & 604.5 & - & 38.8  & 39.2 & 0.0  \\
        Seed-X 17B & 63.4  & 70.1 & 66.5 & 43.0 & 1457.0 & - & 35.6  & - & - \\
        VILA-U 7B & 79.4  & 66.6 & 57.1  & 33.5 & 1401.8 & - & 32.2  & 46.6 & 48.3 \\
        LMFusion 16B & -  & - & 72.1  & - & 1603.7 & 367.8 & 41.7 & 60.0 & -  \\
        Show-o 1.3B & 69.4  & - & - & - & 1097.2 &  - & 27.4 & - & -  \\
        EMU3 8B & 75.1  & 58.5 & 68.2 & 37.2 & 1243.8 & 266.1 & 31.6 & 57.4 & 64.7 \\
        MetaMorph 8B & -  & 75.2 & 71.8  & - & - & - & 41.8 & 58.3 & 60.5   \\
        TokenFlow-XL 14B & 77.6 &  76.8 & 72.6  & 48.2 & 1551.1 & 371.1 & 43.2 & 56.6 & 77.6  \\
        Janus 1.3B & 77.3  & 75.5 & 68.3 & 34.3&1338.0 & - & 30.5  & - & -\\
        Janus Pro 7B & -  & 79.2 & 72.1 & 50.0 & 1567.1 & - & 41.0 & - & - \\
         \rowcolor{green!10}\modelname{} 4B  & 75.9  & 78.6 & 73.8  & 60.1 & 1527.7 & 632.9 & 46.6 & 60.4 & 78.0 \\
        \rowcolor{green!10}\modelname{} 8B  & \textbf{83.1}  & \textbf{83.5} & \textbf{77.5} & \textbf{66.6} & \textbf{1682.6} & \textbf{647.1} & \textbf{50.6} & \textbf{69.0} & \textbf{83.1} \\
        \end{tabular*}
        }
    \end{subtable}%
    \caption{Results on image understanding benchmarks. We highlight the best results in \textbf{bold}. \label{tab:und_results}}
\end{table*}

\paragraph{Image Generation}

\begin{figure}[ht]
  \centering
  \begin{minipage}[b]{0.45\textwidth}
    \centering
    \fontsize{7pt}{8.8pt}\selectfont
    \setlength\tabcolsep{7pt}
    \renewcommand{\arraystretch}{1.1}
    \begin{tabular}{l|ccc}
      Model            & GenEval & DPG-Bench & WISE \\
      \hline
      Chameleon 7B     & 0.39    & –      & -   \\
      Seed-X 17B       & 0.51    & –      & -  \\
      LLaVAFusion 16B  & 0.63    & –      & -  \\
      Show-o 1.3B      & 0.68    & 67.27  & 0.35   \\
      EMU3 8B          & 0.66    & 80.60  & 0.39  \\
      TokenFlow-XL 14B & 0.63    & 73.38  & -   \\
      Janus 1.3B       & 0.61    & 79.68  & 0.18   \\
      Janus Pro 7B     & 0.80    & \textbf{84.19} & 0.35 \\
      \rowcolor{green!10}\modelname{} 4B & 0.81 & 79.36 & 0.50\\
      \rowcolor{green!10}\modelname{} 8B  & \textbf{0.84} & 81.60 & \textbf{0.62} \\
    \end{tabular}
    \captionof{table}{Image generation benchmark results.}
    \label{tab:gen_results}
  \end{minipage}%
  \hfill
  \begin{minipage}[b]{0.5\textwidth}
    \centering
\includegraphics[width=\linewidth]{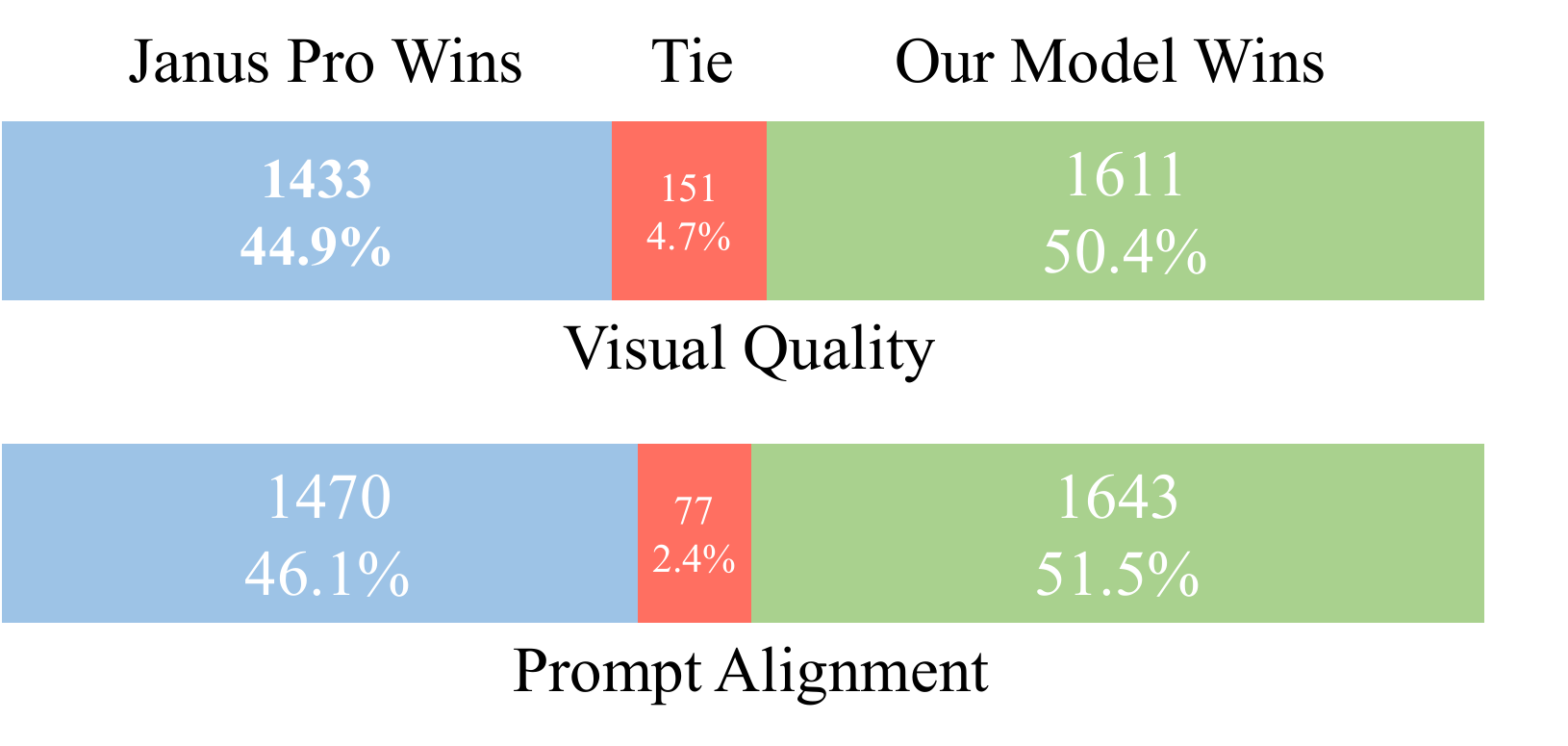}
    \captionof{figure}{Human study results for DPG-Bench between Janus Pro and our model.}
    \label{fig:human_study}
  \end{minipage}
\end{figure}

In the image generation benchmark, we report GenEval~\cite{ghosh2023geneval} and DPG‑Bench~\cite{hu2024ella} to measure prompt alignment, WISE~\cite{niu2025wise} to evaluate world knowledge reasoning capability. As shown in Table~\ref{tab:gen_results}, \modelname{} 8B achieves a GenEval score of 0.84, a WISE score of 0.62, but scores lower on DPG-Bench. Because model-based evaluation for DPG-Bench can be unreliable, we complement these results with a human study on all DPG-Bench prompts in the next section. Furthermore, we also find our instruction tuning dataset \dataname{} yields immediate gains: using only 60k prompt–image pairs, both prompt alignment and visual aesthetics improve markedly, and many generation artifacts are quickly reduced. Although this instruction tuning dataset cannot fully resolve some difficult cases, such as complex human gestures generation, it nonetheless delivers a substantial boost in overall image quality.

\begin{findingbox}[title={Finding 3}]
The model can rapidly adapt to GPT-4o style, enhancing both prompt alignment and visual quality. The model learns more effectively from AI-generated images than from real images.
\end{findingbox}


\subsection{Human Study}

In this section, we conduct a human evaluation comparing \modelname{} 8B and Janus Pro 7B on about 1,000 prompts drawn from the DPG-Bench. For each prompt, annotators compare image pairs side by side on two metrics:

\begin{itemize}
    \item Visual Quality: the instruction is “All images were generated from the same text input using different methods. Please select the BEST image you prefer based on visual appeal, such as layout, clarity, object shapes, and overall cleanliness.”
  \item Prompt Alignment: the instruction is “All images were generated from the same text input using different methods. Please select the image with the BEST image-text content alignment.” 
\end{itemize}

Each metric was assessed in two separate rounds, resulting in roughly 3,000 judgments per criterion. As illustrated in Figure~\ref{fig:human_study}, \modelname{} outperforms Janus Pro on both visual quality and prompt alignment, even though Janus Pro achieves a higher DPG score in Table~\ref{tab:gen_results}. The $p$-values for Visual Quality and Prompt Alignment are 5.05e-06 and 1.16e-05, respectively, indicating that our model significantly outperforms Janus Pro with high statistical confidence.




\section{Future Work}

We are currently extending our unified multimodal to downstream tasks such as image editing, multi‐turn visual dialogue, and interleaved generation. As a first step, we will focus on image reconstruction: feeding images into the image understanding vision encoder and then reconstructing them via the image generation model, to seamlessly bridge image understanding and generation. Building on this capability, we will collect instruction tuning datasets to adapt the model to various downstream applications.

\section{Related Work}

Recent studies have highlighted unified multimodal, capable of both image understanding and generation, as a promising avenue of research. For example, SEED-X~\cite{ge2024seed}, Emu-2~\cite{sun2024generative}, and MetaMorph~\cite{tong2024metamorph} train image features via regression losses, while Chameleon~\cite{team2024chameleon}, Show-o~\cite{xie2024show}, EMU3~\cite{wang2024emu3}, and Janus~\cite{wu2024janus, chen2025janus} adopt an autoregressive discrete token prediction paradigm. In parallel, DreamLLM~\cite{dong2023dreamllm} and Transfusion~\cite{zhou2024transfusion} leverage diffusion objectives for visual generation. To our knowledge, we present the first systematic study of design choice in the autoregressive and diffusion framework.

Regrading to the unifed model training strategy, LMFusion \cite{shi2024llamafusion} builds on the frozen MLLM backbone while incorporating transformer modules for image generation using Transfusion \cite{zhou2024transfusion}. A key similarity between our approach and LMFusion is that both methods freeze the MLLM backbone and train only the image-specific components. However, LMFusion incorporates parallel transformer modules for image diffusion, significantly expanding the model size. In contrast, our method introduces a relatively lightweight diffusion head to enable image generation, maintaining a more manageable overall model size. Concurrent work MetaQuery~\cite{pan2025transfer} also uses learnable queries to bridge frozen pre-trained MLLMs with pre-trained diffusion models, but the diffusion models are in VAE + Flow Matching strategy instead of the more efficient CLIP + Flow Matching one in our \modelname{}.






\section{Conclusion}

In summary, we have presented the first systematic exploration of hybrid autoregressive and diffusion architectures for unified multimodal modeling, evaluating three critical aspects: image representation (CLIP vs. VAE features), training objective (Flow Matching vs. MSE), and training strategy (joint vs. sequential). Our experiments demonstrate that CLIP embeddings paired with a flow matching loss deliver both faster training efficiency and higher quality outputs. Building on these insights, we introduce \modelname{}, a family of state-of-the-art unified models enhanced with a 60k instruction tuning dataset \dataname{} that substantially improves prompt alignment and visual aesthetics. We are actively working on applications for the unified model, including iterative image editing, visual dialogue, and step-by-step visual reasoning.


\appendix


\section{Prompt used in Figure~\ref{fig:human_fig}}

\begin{itemize}[itemsep=0.6em, topsep=1em]
  \item A blue BMW parked in front of a yellow brick wall.
  \item A woman twirling in a sunlit alley lined with colorful walls, her summer dress catching the light mid-spin.
  \item A group of friends having a picnic.
  \item A lush tropical waterfall, `Deep Learning` on a reflective metal road sign.
  \item A blue jay standing on a large basket of rainbow macarons.
  \item A sea turtle swimming above a coral reef.
  \item A young woman with freckles wearing a straw hat, standing in a golden wheat field.
  \item Three people.
  \item A man talking animatedly on the phone, his mouth moving rapidly.
  \item A wildflower meadow at sunrise, `BLIP3o` projected onto a misty surface.
  \item A rainbow-colored ice cavern, `Salesforce` drawn in the wet sand.
  \item A giant glass bottle filled with a miniature summer forest inside.
  \item Walk through of frozen streets of Manhattan, New York City—frozen trees and a frozen Empire State Building.
  \item A lighthouse standing alone in a stormy sea
  \item A lone wolf beneath shimmering northern lights.
  \item A glowing deer walking through a neon-lit futuristic jungle.
  \item A couple walking hand in hand through a vibrant autumn park, leaves gently falling around them.
  \item A curious vessel, shaped like a giant green broccoli, floating on a sparkling ocean under bright sunlight.
  \item `Transformer` written on the road.
  \item `Diffusion` on the blue T-shirt.
  \item A golden retriever lying peacefully on a wooden porch, with autumn leaves scattered around.
  \item A raccoon wearing a detective’s hat, solving mysteries with a magnifying glass.
  \item A cyberpunk woman with glowing tattoos and a mechanical arm beneath a holographic sky.
  \item The reflection of a snowy mountain peak in a crystal-clear alpine lake, forming a perfect mirror image.
  \item A man sipping coffee on a sunny balcony filled with potted plants, wearing linen clothes and sunglasses, basking in the morning light.
\end{itemize}

\newpage

\bibliography{main}
\bibliographystyle{plain}
\end{document}